\documentclass[11pt]{article}
\usepackage[T1]{fontenc}
\usepackage[utf8]{inputenc}
\usepackage{lmodern}
\usepackage[margin=1in]{geometry}
\usepackage{amsmath,amssymb,amsthm,mathtools,bm}
\usepackage{booktabs,array,enumitem,microtype}
\usepackage{xcolor}
\usepackage[colorlinks=true,linkcolor=blue!45!black,citecolor=blue!45!black,urlcolor=blue!45!black]{hyperref}
\newtheorem{theorem}{Theorem}[section]
\newtheorem{lemma}[theorem]{Lemma}
\newtheorem{proposition}[theorem]{Proposition}
\newtheorem{corollary}[theorem]{Corollary}
\newtheorem{assumption}[theorem]{Assumption}
\theoremstyle{remark}

\newcommand{\E}{\mathbb E}
\newcommand{\Pbb}{\mathbb P}
\newcommand{\R}{\mathbb R}
\newcommand{\eps}{\varepsilon}
\newcommand{\norm}[1]{\left\lVert#1\right\rVert}
\newcommand{\abs}[1]{\left\lvert#1\right\rvert}
\newcommand{\clip}{\operatorname{clip}}
\newcommand{\range}{\operatorname{range}}
\newcommand{\tr}{\operatorname{tr}}

\newcommand{\cA}{\mathcal A}
\newcommand{\cS}{\mathcal S}
\newcommand{\cK}{\mathcal K}

\newcommand{\argmax}{\operatorname*{arg\,max}}

\setlist{itemsep=3pt,topsep=5pt}
\hypersetup{pdftitle={Computationally Efficient Offline Policy Optimization with All-Policy Linear Realizability},pdfauthor={Zhijie Zheng},pdfsubject={Offline reinforcement learning, linear realizability, trajectory data, computational complexity}}
\title{Computationally Efficient Offline Policy Optimization\\with All-Policy Linear Realizability}
\author{Zhijie Zheng\\Peking University\\\texttt{2501210512@stu.pku.edu.cn}}
\date{}

\begin{document}
\maketitle
\begin{abstract}
We study offline policy optimization in finite-horizon Markov decision processes where every deterministic Markov policy has an action-value function linear in known features, with coefficient norm at most $B$. From independent complete trajectories of one unknown Markov behavior policy, and under all-policy occupancy coverage, we give an algorithm with $\widetilde O(\eps^{-2})$ sample dependence and computational complexity polynomial in $\log(2+B)$. The algorithm performs exactly one convex regression per layer. Its stopped-return targets are the values of a single consistent Markov witness policy. A shared reference design and bounded prediction constraints convert fast clipped-regression error into pointwise confidence for the raw linear predictions. Truncated leverage controls the performance loss without requiring empirical covariance approximation at scale $B^{-2}$. Rounding confidence radii upward preserves the exact witness and yields a polynomial-time bit implementation for rational numerical inputs. The guarantees allow any fixed initial-state distribution; realizability and coverage need only be checked for deterministic policies. We also establish a covariance-coverage variant with uniform approximate realizability and polynomial dependence on $B$. A two-action lower bound shows that the accuracy exponent $2$ is necessary.
\end{abstract}

\section{Introduction}
The action-value functions of all policies can lie in a low-dimensional linear class without the transition model being linear or arbitrary Bellman backups remaining in that class. Consequently, ordinary fitted value iteration need not have realizable regression targets even under all-policy linear value realizability. Offline learning must address this issue using only the trajectories already collected.

Our algorithm uses stopped returns and a carefully chosen continuation value. At a state where the fitted action range exceeds twice its uncertainty, a pessimistic query lies inside the range of the true action-value function. A mixture of true extreme actions realizes this query exactly. Elsewhere the analytical policy follows the behavior policy. Extending this policy backward makes every subsequent stopped-return target the action-value function of the same consistent suffix policy. This witness is used only in the proof; neither its actions nor the behavior probabilities are needed by the learner.

The remaining challenge is certification when the coefficient bound $B$ is large. A direct relative-covariance argument at regularization scale $B^{-2}$ incurs a numerical $B^2$ sample cost. We avoid that step. One reference sample is used both to constrain raw predictions at observed features and to define a regularized confidence geometry. Fast regression controls clipped squared error. A uniform two-coefficient reference bound transfers this control to the adaptively selected witness and fitted coefficients, and the prediction constraints recover a raw ellipsoidal confidence bound. Only truncated leverage is needed to evaluate the final policy, and its bound depends on $B$ through a logarithmic determinant.

The algorithm uses one backward pass with one convex regression per layer. Its sample bound has accuracy exponent $2$, while its sample count, training computation, output size, and deployment cost depend polynomially on $\log(2+B)$. For rational numerical inputs, rounding each confidence radius upward makes the stopped labels rational while preserving the exact witness. A common coefficient grid controls encoding lengths throughout the backward pass and gives a polynomial-time bit implementation. The guarantees hold for any fixed initial-state distribution, and deterministic-policy realizability and coverage suffice.

\paragraph{Related work.}
Weisz, Gy\"orgy, and Szepesv\'ari~\cite{weisz2023} introduce state skipping for statistically efficient online learning under all-policy linear value realizability. Their admissible-realizability argument uses randomized choices between actions with high and low values. In the online setting, Mhammedi~\cite{mhammedi2025} obtains polynomial sample and oracle complexity using a cost-sensitive classification oracle. Tkachuk, Weisz, and Szepesv\'ari~\cite{tkachuk2024} show that complete trajectories suffice for statistically efficient offline policy optimization under linear $q^\pi$-realizability and concentrability. Their sample bound already has quadratic accuracy dependence and logarithmic dependence on the coefficient radius, while computational efficiency remains open in their setting. Tkachuk, Szepesv\'ari, and Tan~\cite{tkachuk2026} study trajectory-based policy evaluation when the behavior policy is known and improve the statistical analysis of policy optimization. Our contribution is a computationally efficient construction under the assumptions below, with one regression per layer and an explicit finite-bit implementation. The analysis combines a single consistent witness with shared-design certification. We use standard squared-loss fast-rate, matrix concentration, and convex separation tools~\cite{grunwald2021,tropp2015,gls1988,bental2001} and provide the quantitative bounds needed here.

\section{Model and main result}\label{sec:model}
Consider a layered Markov decision process with horizon $H\ge1$, a fixed initial-state distribution $\zeta$, measurable state spaces $\cS_h$, a finite nonempty action set $\cA$, and rewards in $[0,1]$. The distribution $\zeta$ need not be known. Policies are state-dependent, nonstationary Markov policies and may be randomized. All action choices use fixed measurable tie breaking. Write $L_h=H-h+1$, let $Q_h^\pi,V_h^\pi$ be the usual undiscounted values, and define $V^\pi=\E_{S_1\sim\zeta}V_1^\pi(S_1)$ and $V^*=\sup_\pi V^\pi$. All occupancy measures use the same initial distribution $\zeta$.

Known measurable features $\phi_h:\cS_h\times\cA\to\R^d$ satisfy $d\ge1$ and $\norm{\phi_h(s,a)}_2\le1$. The action set can be enumerated and features can be evaluated; these costs are counted. The arithmetic model permits arithmetic, comparisons, and square roots. No enumeration or finite-cardinality assumption is imposed on the state spaces. Values and stopped-return means at zero-probability state-action pairs are defined by forcing the pair and then following the Markov kernels, rather than by choosing a version of a conditional expectation on a null event.

\begin{assumption}[All-policy linear realizability]\label{ass:linear}
A known $B\ge1$ satisfies: for every deterministic Markov policy $\pi$ and layer $h$, there exists $\theta_h^\pi\in\R^d$ with
\[
Q_h^\pi(s,a)=\phi_h(s,a)^\top\theta_h^\pi
\quad\text{for all }(s,a),\qquad \norm{\theta_h^\pi}_2\le B.
\]
\end{assumption}
Lemma~\ref{lem:randomization} extends this assumption to randomized Markov policies, including the state-dependent mixtures used in the proof. Bellman completeness is not required.

Data are independent complete trajectories from one fixed, unknown Markov behavior policy $b$. Let $\mu_h=d_h^b$ be its state-action occupancy and $\mu_h^S$ its state marginal. Layers within a trajectory may be dependent. The algorithm does not require $b_h(a\mid s)$, but does require complete logged suffixes.
\begin{assumption}[All-policy occupancy coverage]\label{ass:coverage}
A known $C\ge1$ satisfies $d_h^\pi\le C\mu_h$ as measures, for every deterministic Markov policy $\pi$ and every layer $h$.
\end{assumption}

\begin{lemma}[Closure under policy randomization]\label{lem:randomization}
Fix an initial-state distribution and let $\eta\ge0$. Suppose that, for every deterministic Markov policy $\pi$ and layer $h$, there is a vector $\theta_h^\pi$ with $\|\theta_h^\pi\|_2\le B$ such that
\[
\sup_{s,a}|Q_h^\pi(s,a)-\phi_h(s,a)^\top\theta_h^\pi|\le\eta.
\]
Then the same statement holds for every randomized Markov policy, with the same $B$ and $\eta$. Likewise, either coverage condition
\[
d_h^\pi\le C\mu_h,\qquad
\E_{d_h^\pi}[\phi_h\phi_h^\top]\preceq C\E_{\mu_h}[\phi_h\phi_h^\top]
\]
extends from all deterministic Markov policies to all randomized Markov policies without changing $C$.
\end{lemma}
\begin{proof}
Enumerate the finite action set. Implement a randomized Markov policy by inverse cumulative probabilities using independent variables $U_1,\ldots,U_H$, uniform on $(0,1)$. Fixing $U=u$ defines a measurable deterministic Markov policy $\pi^u$. Integrating over $u$ recovers the original trajectory law and, for every forced state-action pair, its continuation law. Thus
\[
d_h^\pi=\E_U d_h^{\pi^U},\qquad
Q_h^\pi(s,a)=\E_U Q_h^{\pi^U}(s,a).
\]
The second identity holds pointwise because the pair is forced, rather than obtained by conditioning the random seeds on a visit.

Fix a layer and a finite set $F$ of state-action pairs. Let $\Phi_F$ be the matrix of their feature rows and $Q_F^\pi$ the corresponding value vector. Each $Q_F^{\pi^u}$ belongs to the compact convex set
\[
\Phi_F B_2(0,B)+[-\eta,\eta]^{|F|},
\]
so its expectation $Q_F^\pi$ also belongs to this set. The closed constraints
$|Q_h^\pi(s,a)-\phi_h(s,a)^\top\theta|\le\eta$
therefore have a common solution in $B_2(0,B)$ for every finite collection of pairs. Compactness and the finite intersection property give one coefficient vector satisfying all pairs simultaneously. No measurable selection of coefficients for $\pi^u$ is needed. Integrating the deterministic occupancy or covariance inequalities proves the coverage assertions.
\end{proof}

\begin{theorem}[Quadratic accuracy and logarithmic radius dependence]\label{thm:main}
Under Assumptions~\ref{ass:linear} and~\ref{ass:coverage}, for $0<\eps\le1$ and $0<\delta<1$, the algorithm in Section~\ref{sec:algorithm} returns a deterministic Markov policy $\widehat\pi$ with
\[
\Pbb\{V^*-V^{\widehat\pi}\le27\eps/32\}\ge1-\delta.
\]
The algorithm performs exactly $H$ convex regressions and uses the explicit number $N=n+Hm_{\rm fit}$ of complete trajectories in \eqref{eq:budget}. This budget is polynomial in
\begin{equation}\label{eq:parameters}
H,\ d,\ C,\ \eps^{-1},\ \log(2+B),\ \log(1/\delta),
\end{equation}
and has $\widetilde O(\eps^{-2})$ dependence when $H,d,C,B,\delta$ are fixed. Its arithmetic computation is polynomial in \eqref{eq:parameters} and $|\cA|$.

For rationally encoded parameters, observed rewards, and evaluated features, it has a bit implementation polynomial in those encoding lengths, the feature and state-processing costs, and the displayed parameters. The output representation and policy evaluation satisfy corresponding polynomial bounds. In particular, this implementation does not charge polynomially in the numerical value of $B$.
\end{theorem}
Appendix~\ref{comp:bits} specifies the finite-input model. The lower bound in Appendix~\ref{sw:lower-sec} shows that the accuracy exponent $2$ is necessary. The dependence on horizon, dimension, and coverage remains a subject for further improvement.

\section{Algorithm: a shared design and one backward pass}\label{sec:algorithm}
All sample sizes and numerical tolerances are chosen before inspecting the data. Throughout, $\log$ is natural and $\log_2$ is base two.

\subsection{Reference design and data allocation}
Set $\lambda_0=H^2/(4B^2)$. For an integer $n\ge1$, define
\begin{align}
A_n&=1+2d\left\lceil\log_2\!\left(1+\frac{4nB}{H}\right)\right\rceil
       +\left\lceil\log_2\frac{8H}{\delta}\right\rceil,\label{eq:An}\\
J_n&=4d\left\lceil\log_2\!\left(1+\frac n{d\lambda_0}\right)\right\rceil
       +2\left\lceil\log_2\frac{8H}{\delta}\right\rceil,\label{eq:Jn}\\
q&=\min\left\{1,\frac{A_n}{n}\right\},&
s_n&=\frac H{\lceil\sqrt n\rceil},&R_n^2&=128H^2A_n.\label{eq:statparams}
\end{align}
Choose $n$ to be the first power of two satisfying
\begin{equation}\label{eq:nchoice}
v_n^2:=\frac{CR_n^2J_n}{n}
       =\frac{128CH^2A_nJ_n}{n}\le\frac{\eps^2}{64H^2}.
\end{equation}
Appendix~\ref{comp:budget} gives an explicit upper bound on this deterministic doubling search.

From $n$ reference trajectories form, separately at every layer,
\begin{equation}\label{eq:reference}
x_{ih}=\phi_h(S_h^i,A_h^i),\qquad
V_h=\lambda_0I+\sum_{i=1}^n x_{ih}x_{ih}^\top.
\end{equation}
These are unnormalized design matrices. Define the theoretical confidence functions
\begin{equation}\label{eq:uncertainty}
u_h(s,a)=R_n\sqrt{\phi_h(s,a)^\top V_h^{-1}\phi_h(s,a)},\qquad
U_h(s)=\max_a u_h(s,a),\qquad \bar U_h(s)=\min\{H,U_h(s)\}.
\end{equation}
The algorithm uses their upper rounded version
\begin{equation}\label{eq:rounding}
\kappa=\frac{\eps}{64H},\qquad
W_h(s)=\kappa\left\lceil\frac{U_h(s)}\kappa\right\rceil,
\qquad U_h(s)\le W_h(s)<U_h(s)+\kappa.
\end{equation}
With rational input, $W_h$ is computed by exact comparisons of rational squares. No irrational square root is inserted into a label.

For the fits define integer logarithmic budgets and block sizes by
\begin{align}
F_n&=d\left\lceil\log_2\!\left(1+\frac{128BH}{q}\right)\right\rceil
          +\left\lceil\log_2\frac{8H}{\delta}\right\rceil,
&m_{\rm fit}&=\left\lceil\frac{64H^2F_n}{q}\right\rceil,\label{eq:mfit}\\
N&=n+Hm_{\rm fit}.\label{eq:budget}
\end{align}
Use one independent fitting block of $m_{\rm fit}$ complete trajectories for each layer. All $H+1$ blocks are mutually independent. No validation sample is required.

\subsection{Backward regression and interval queries}
At layer $h$ use the convex domain
\begin{equation}\label{eq:feasible}
\cK_h=\left\{w\in\R^d:\norm w_2\le B,\quad
-s_n\le x_{ih}^\top w\le L_h+s_n\quad(1\le i\le n)\right\}.
\end{equation}
There are only $2n$ reference inequalities, for the actions actually recorded in the reference trajectories. Process layers $h=H,H-1,\ldots,1$ as follows.
\begin{enumerate}[leftmargin=*,label=\textbf{\arabic*.}]
\item \textbf{Construct stopped labels.} Later-layer functions have already been frozen. On each trajectory in the current fitting block, compute the suffix recursion
\begin{equation}\label{eq:labels}
Y_H=R_H,\qquad
Y_t=R_t+(1-\omega_{t+1}(S_{t+1}))v_{t+1}(S_{t+1})
             +\omega_{t+1}(S_{t+1})Y_{t+1},
\end{equation}
for $t=H-1,\ldots,h$. A value $\omega=1$ means continuation along the logged behavior suffix; $\omega=0$ means substitution of the bounded query $v$.
\item \textbf{Fit once.} Compute $\widehat\theta_h\in\cK_h$ whose empirical raw squared loss on $(\phi_h(S_h,A_h),Y_h)$ is within $q/8$ of the minimum over $\cK_h$. The tolerance includes numerical rounding. Set $\widehat q_h(s,a)=\phi_h(s,a)^\top\widehat\theta_h$.
\item \textbf{Freeze the rule and the query.} At every state define
\begin{align}
\omega_h(s)&=\begin{cases}
0,&\range_a\widehat q_h(s,a)\ge2W_h(s),\\
1,&\text{otherwise},
\end{cases}\label{eq:skip}\\
v_h(s)&=\clip_{[0,L_h]}\!\left(\max_a\widehat q_h(s,a)-W_h(s)\right).
\label{eq:query}
\end{align}
The range is the maximum minus the minimum over $\cA$; the query is used only at states with $\omega_h=0$.
\end{enumerate}
Return the deterministic policy
\begin{equation}\label{eq:output}
\widehat\pi_h(s)\in\argmax_a\widehat q_h(s,a).
\end{equation}
This is one backward pass. Each layer uses a fresh block even though its labels are computed from a complete suffix. The reference block is reused at every stage. The domain contains zero, so each regression is feasible on every input. Clipping ensures $0\le Y_h\le L_h$ on every history, including a history on which a statistical confidence event fails. At deployment, only the $H$ final coefficient vectors and feature evaluations are required.

\section{One exact witness for all stopped targets}\label{sec:witness}
The essential invariant concerns a true Markov policy, not a propagated approximation to a Bellman backup. We first state the interval observation behind it.
\begin{lemma}[Interval queries]\label{lem:interval}
Let $q,Q$ be two real vectors on a finite action set and suppose $|q(a)-Q(a)|\le W$ for every action. If $\range_a q(a)\ge2W$, then
\[
\max_a q(a)-W\in[\min_a Q(a),\max_a Q(a)].
\]
Every point of this interval is the expectation of $Q$ under a mixture of its true minimizing and maximizing actions.
\end{lemma}
\begin{proof}
For a maximizing action of $q$, $\max q-W\le Q(a)\le\max Q$. For the other endpoint,
$\max q-W\ge\min q+W\ge\min Q$.
If the endpoints differ, assign probability $(v-\min Q)/(\max Q-\min Q)$ to a maximizing action to realize the value $v$; if they coincide, any action suffices. Fixed tie breaking gives measurable statewise mixtures.
\end{proof}

\begin{lemma}[Consistent witness and exact stopped targets]\label{lem:witness}
Consider a backward pass using \eqref{eq:labels}, \eqref{eq:skip}, and \eqref{eq:query}, with any nonnegative statewise confidence functions $W_h$. Suppose that, whenever the suffix witness has been constructed after layer $h$, the next fit satisfies
\begin{equation}\label{eq:witness-confidence}
|\widehat q_h(s,a)-Q_h^{\pi_{>h}}(s,a)|\le W_h(s)
\quad\text{at every }(s,a),
\end{equation}
where $\pi_{>h}$ is that fixed suffix, extended arbitrarily before it. Then the construction yields one full Markov policy $\pi_{\rm w}$ such that
\begin{equation}\label{eq:witness-invariant}
V_h^{\pi_{\rm w}}(s)=v_h(s)\quad\text{if }\omega_h(s)=0,
\qquad \pi_{{\rm w},h}(\cdot\mid s)=b_h(\cdot\mid s)
\quad\text{if }\omega_h(s)=1.
\end{equation}
Before each fit, its stopped-label conditional mean is exactly
\begin{equation}\label{eq:target}
F_h(s,a):=\E_b[Y_h\mid(s,a)\text{ forced at }h]
       =Q_h^{\pi_{>h}}(s,a).
\end{equation}
Earlier extensions leave all established suffix values and confidence inequalities unchanged. The witness need not be computed by the learner.
\end{lemma}
\begin{proof}
At $H$ the empty suffix is the base case, and the mean label is the mean immediate reward, independent of any continuation policy. Suppose the suffix has been constructed and satisfies \eqref{eq:witness-invariant} at all later layers. A stopped return follows $b$ until the first later non-skipped state and substitutes $v_t$ there. Before stopping the suffix witness follows $b$, and at stopping its state value is exactly $v_t$. If no stop occurs, the witness follows $b$ to the horizon. Backward conditioning through the Markov kernels proves \eqref{eq:target}, at every forced pair.

By \eqref{eq:witness-confidence} and Lemma~\ref{lem:interval}, at a non-skipped state the untruncated query $\max_a\widehat q_h-W_h$ lies in the true action-value interval of the existing suffix. This interval is a subset of $[0,L_h]$, so clipping leaves the query unchanged. Choose a mixture of the true extreme actions which realizes it exactly. At a skipped state choose $b_h$. This defines the stage-$h$ action of the witness without changing its suffix or its stage-$h$ action values. Continuing backward gives one full policy. The use of a rational upper bound $W_h\ge U_h$ only enlarges the confidence margin; the interval argument and the witness equalities remain exact.
\end{proof}

\section{Certification and policy quality}\label{sec:proof}
\subsection{Shared reference certification without covariance approximation}
There are two uniform reference events. For coefficients $w,\theta$, write
\[
g_{w,\theta,h}(x)=
 \bigl(\clip_{[0,L_h]}(x^\top w)-\clip_{[0,L_h]}(x^\top\theta)\bigr)^2.
\]
Lemma~\ref{stat:reference} states that, simultaneously over layers and all pairs with $\norm w_2,\norm\theta_2\le B$,
\begin{equation}\label{eq:ref-bound}
P_n g_{w,\theta,h}\le2P g_{w,\theta,h}+\frac{32H^2A_n}{n},
\end{equation}
where $P=\mu_h$ and $P_n$ is the layer-$h$ reference average.
Lemma~\ref{stat:leverage} gives, simultaneously over layers,
\begin{equation}\label{eq:leverage}
\E_{\mu_h}\min\{1,\phi_h^\top V_h^{-1}\phi_h\}\le J_n/n.
\end{equation}
Each event fails with probability at most $\delta/8$. The first event is uniform in both coefficients, allowing both a fitted vector and the coefficient of a data-dependent witness to be substituted. The second concerns the final reference matrix, not an independent or pre-observation matrix.

Condition on the reference block and successful fits at layers $H,\ldots,h+1$. The suffix witness is fixed. By Lemma~\ref{lem:witness}, the conditional regression mean is its true $Q_h$. Assumption~\ref{ass:linear} supplies a coefficient $\theta_h^*$ of norm at most $B$, and $0\le Q_h\le L_h$ makes this coefficient feasible for every reference constraint. A fresh independent fitting block and the fast squared-loss bound in Lemma~\ref{stat:fast} give
\begin{equation}\label{eq:fitrisk}
\E_{\mu_h}\bigl(\clip_{[0,L_h]}(\phi_h^\top\widehat\theta_h)-Q_h^{\pi_{>h}}\bigr)^2\le q
\end{equation}
with conditional failure probability at most $\delta/(8H)$. There is no propagation factor in \eqref{eq:fitrisk}: its comparator is the exact current witness value.

For this fitted vector, \eqref{eq:ref-bound} and $q\le A_n/n$ bound the mean squared clipped error on the reference sample by $34H^2A_n/n$. Feasibility places every raw reference prediction within $s_n$ of $[0,L_h]$. Therefore, as detailed in Lemma~\ref{stat:confidence},
\begin{align}
\norm{\widehat\theta_h-\theta_h^*}_{V_h}^2
&\le\lambda_0(2B)^2+68H^2A_n+2ns_n^2\notag\\
&\le71H^2A_n<R_n^2.\label{eq:rawconfidence}
\end{align}
Matrix Cauchy--Schwarz implies
\[
|\widehat q_h(s,a)-Q_h^{\pi_{>h}}(s,a)|
\le u_h(s,a)\le U_h(s)\le W_h(s)
\]
at every state-action pair. Lemma~\ref{lem:witness} now extends the witness exactly. This closes the induction: a certified future suffix makes the current target realizable, and the current fit supplies the confidence needed for its extension.

A first-failure union bound over the $H$ regression stages, together with the two reference events, costs at most $3\delta/8\le\delta$. It conditions a stage only on its already exposed successful history, not on the eventual success of all fits. All remaining statements are made on this simultaneous good event.

\subsection{Truncation is sufficient for the performance bound}
For any reachable state marginal $\nu_h^S$, preserve a policy prefix giving that marginal and choose at layer $h$ an action maximizing the leverage in \eqref{eq:uncertainty}. This is an admissible Markov policy. Coverage and \eqref{eq:leverage}, using $R_n^2\ge H^2$, give
\begin{equation}\label{eq:uncertainty-moment}
\E_{\nu_h^S}\bar U_h^2
\le CR_n^2\E_{\mu_h}\min\{1,\phi_h^\top V_h^{-1}\phi_h\}
\le v_n^2.
\end{equation}
The assertion is uniform over reachable state marginals, including those of the fitted and witness policies. Hence
\begin{equation}\label{eq:rounded-moment}
\E_{\nu_h^S}\min\{H,W_h\}\le\E_{\nu_h^S}\bar U_h+\kappa\le v_n+\kappa.
\end{equation}
Untruncated uncertainty may be large on rare states. It does not need a small average or a uniform small upper bound.

\begin{lemma}[Performance from a certified witness]\label{lem:performance}
For any backward pass satisfying Lemma~\ref{lem:witness}, let $\widehat\pi$ be raw greedy as in \eqref{eq:output}. If a number $a$ satisfies
\[
\E_{\nu_h^S}\min\{H,W_h(S)\}\le a
\quad\text{for all reachable state marginals and all }h,
\]
then $V^*-V^{\widehat\pi}\le6Ha$.
\end{lemma}
\begin{proof}
At a non-skipped state, the exact witness value is $\max_a\widehat q_h-W_h$, so its suboptimality relative to the best action under its own continuation is at most $2W_h$. At a skipped state, the predicted range is less than $2W_h$, and confidence enlarges the true range by at most $2W_h$; the witness averages those true values under $b_h$. All true action values lie in $[0,L_h]$. Thus, at every state,
\begin{align*}
\max_a Q_h^{\pi_{\rm w}}(s,a)-V_h^{\pi_{\rm w}}(s)
&\le\min\{H,4W_h(s)\}\le4\min\{H,W_h(s)\},\\
\max_a Q_h^{\pi_{\rm w}}(s,a)-Q_h^{\pi_{\rm w}}(s,\widehat\pi_h(s))
&\le\min\{H,2W_h(s)\}\le2\min\{H,W_h(s)\}.
\end{align*}
The second inequality follows because $\widehat\pi$ maximizes the raw fitted prediction. For any policies $p,q$, telescoping the values along a trajectory of $p$ gives
\[
V^p-V^q=\sum_{h=1}^H\E_{d_h^{p,S}}
 \left[\E_{A\sim p_h(\cdot\mid S)}Q_h^q(S,A)-V_h^q(S)\right].
\]
Apply this first with $p$ optimal and $q=\pi_{\rm w}$, and then with $p=\widehat\pi$ and $q=\pi_{\rm w}$. The displayed advantages bound $V^*-V^{\pi_{\rm w}}$ by $4Ha$ and $V^{\pi_{\rm w}}-V^{\widehat\pi}$ by $2Ha$. Adding gives the claim.
\end{proof}

Combining Lemma~\ref{lem:performance} with \eqref{eq:rounded-moment} proves
\begin{equation}\label{eq:performance}
V^*-V^{\widehat\pi}\le6H(v_n+\kappa)
\le6H\left(\frac{\eps}{8H}+\frac{\eps}{64H}\right)
=\frac{27\eps}{32}<\eps.
\end{equation}
The sample and computational assertions of Theorem~\ref{thm:main} follow from Appendix~\ref{comp:budget} and Appendix~\ref{comp:bits}.

\section{Covariance coverage and approximate realizability}\label{sec:scope}
The main theorem uses occupancy coverage to transfer a truncated reference-design uncertainty, yielding logarithmic coefficient-radius dependence in both sampling and computation.

Appendix~\ref{sw:appendix} gives an extension under the weaker condition
$\E_{d_h^\pi}[\phi_h\phi_h^\top]\preceq C\E_{\mu_h}[\phi_h\phi_h^\top]$.
For a known uniform realizability error
$\bar\eta\le\eps/(64H\sqrt{Cd})$, it retains $H$ regressions with
\[
\widetilde O\!\left(CH^3d(B^2+H^2d)\eps^{-2}\right)
\]
trajectories. For an arbitrary known approximation bound, Corollary~\ref{sw:arbitrary-eta} gives the additive guarantee
$V^*-V^{\widehat\pi}\le3\eps/4+54H\sqrt{Cd}\,\bar\eta$
with the same sample upper bound. The witness and policy-quality arguments are the same lemmas proved above; only regression certification and uncertainty transfer change. This variant has polynomial dependence on $B$. Whether logarithmic radius dependence can be retained under covariance coverage and approximate realizability remains open.

Both results apply to complete trajectories from a single Markov behavior policy and use known features with explicit evaluation access. Realizability is pointwise over all state-action pairs. Lemma~\ref{lem:randomization} shows that deterministic policies suffice in both the realizability and coverage assumptions, including for the approximate variant.

\appendix

\section{Statistical guarantees for the shared reference design}
\label{stat:appendix}

We use the reference matrices and parameters defined in
\eqref{eq:reference} and \eqref{eq:statparams}. The matrices are
unnormalized. For a function $f$, $Pf$ and $P_mf$ denote its population
and empirical averages, respectively. For a fresh regression block, the
population distribution is conditional on the reference block and all
already processed later-layer blocks. Within a fixed layer, reference
observations are i.i.d. with state--action distribution $\mu_h$; no
independence between layers of a single trajectory is needed. All
logarithms in this appendix are natural unless a base is specified, and
$\norm{z}_V^2=z^{\top}Vz$.

\subsection{Fast regression for clipped predictions}

\begin{lemma}[Fast clipped regression]
\label{stat:fast}
Condition on a sigma-field $\mathcal G$. Let $(X_i,Y_i)_{i=1}^m$ be
conditionally independent copies of $(X,Y)$ with
\[
 \norm{X}_2\le1,\quad 0\le Y\le L_h,\quad
 \E[Y\mid X,\mathcal G]=F(X)=X^{\top}\theta^*,\quad
 \norm{\theta^*}_2\le B.
\]
Suppose $\theta^*$ belongs to a $\mathcal G$-measurable feasible set
$\cK\subseteq\{w:\norm{w}_2\le B\}$, and a feasible output
$\widehat w$ satisfies
\[
 P_m(X^{\top}\widehat w-Y)^2
 \le\inf_{w\in\cK}P_m(X^{\top}w-Y)^2+q/8.
\]
If
\[
 m\ge\frac{64H^2}{q}
 \left[d\log\left(1+\frac{128BH}{q}\right)+\log\frac1\rho\right],
 \qquad 0<\rho<1,
\]
then, with conditional probability at least $1-\rho$,
\[
 P\bigl(\clip_{[0,L_h]}(X^{\top}\widehat w)-F(X)\bigr)^2\le q.
\]
The guarantee is simultaneous over all outputs satisfying the displayed
empirical optimization bound.
\end{lemma}

\begin{proof}
Work conditionally on $\mathcal G$. Put
\[
 f_w=\clip_{[0,L_h]}(X^{\top}w),\qquad
 \Delta_w=(f_w-Y)^2-(F-Y)^2,\qquad r(w)=P(f_w-F)^2.
\]
The conditional mean property gives $P\Delta_w=r(w)$. All three of
$f_w,F,Y$ lie in $[0,L_h]$, so
\[
 \Delta_w=(f_w-F)(f_w+F-2Y),\quad
 P\Delta_w^2\le4H^2r(w),\quad
 \abs{\Delta_w}\le H^2.
\]
In particular, $\abs{\Delta_w-P\Delta_w}\le2H^2$. A one-sided
Bernstein inequality and $\sqrt{8ab}\le a/2+4b$ imply, for a fixed $w$,
\begin{align*}
 r(w)-P_m\Delta_w
 &\le\sqrt{8H^2r(w)t/m}+4H^2t/(3m)\\
 &\le r(w)/2+16H^2t/(3m)
\end{align*}
except with probability $e^{-t}$. Enlarging the numerical constant gives
\begin{equation}
 r(w)\le2P_m\Delta_w+12H^2t/m.
 \label{stat:fixed-fast}
\end{equation}

Take a Euclidean $q/(64H)$-net of the radius-$B$ ball, with centers in
the ball and cardinality at most $(1+128BH/q)^d$. The usual volume
argument for a maximal separated subset proves this cardinality bound.
With
$t=d\log(1+128BH/q)+\log(1/\rho)$, a union bound makes
\eqref{stat:fixed-fast} valid on the net. Clipping is one-Lipschitz, and
squared loss for predictions and labels in $[0,L_h]$ is $2H$-Lipschitz
in the prediction. For a nearest net point $w'$, it follows that
\[
 \abs{r(w)-r(w')}\le q/32,\qquad
 \abs{P_m\Delta_w-P_m\Delta_{w'}}\le q/32.
\]
Consequently, on the same event, every $\norm{w}_2\le B$ obeys
\[
 r(w)\le2P_m\Delta_w+3q/32+12H^2t/m.
\]
Clipping cannot increase squared loss against a label in $[0,L_h]$.
Feasibility of $\theta^*$ and the optimization guarantee therefore give
\[
 P_m(f_{\widehat w}-Y)^2
 \le P_m(X^{\top}\widehat w-Y)^2
 \le P_m(F-Y)^2+q/8.
\]
Thus $r(\widehat w)\le q/4+3q/32+3q/16=17q/32\le q$.
The concentration event covers the entire coefficient ball, so the
random feasible set creates no additional uniformity requirement.
\end{proof}

In the backward algorithm, the current target is the action-value
function of the already constructed witness suffix. It is exactly linear
with coefficient norm at most $B$ and lies in $[0,L_h]$. Hence its
coefficient satisfies every constraint in \eqref{eq:feasible}. If the
mean identity is first stated conditional on $(S_h,A_h)$, conditioning
again on $X=\phi_h(S_h,A_h)$ gives the identity in the lemma by the
tower property. Fresh complete trajectories give the required conditional
independence. Taking $\rho=\delta/(8H)$ yields the stated fitting budget.
There is no propagation loss here: the conditional regression mean is
already the true action-value function against which confidence is needed.

\subsection{Uniform squared error on the reference block}

\begin{lemma}[Uniform squared reference bound]
\label{stat:reference}
With probability at least $1-\delta/8$ over the reference trajectories,
simultaneously for all layers and all $\norm{w}_2,\norm{\theta}_2\le B$,
\begin{equation}
 \frac1n\sum_{i=1}^n g_{w,\theta,h}(x_{ih})
 \le2\E_{\mu_h}g_{w,\theta,h}(\phi_h(S,A))
                     +\frac{32H^2A_n}{n},
 \label{stat:reference-event}
\end{equation}
where
\[
 g_{w,\theta,h}(x)=
 \bigl(\clip_{[0,L_h]}(x^{\top}w)
          -\clip_{[0,L_h]}(x^{\top}\theta)\bigr)^2.
\]
The substituted coefficients may depend on both the reference block and
all fitting blocks.
\end{lemma}

\begin{proof}
For a fixed pair, $0\le g\le H^2$ and
$\operatorname{Var}(g)\le Pg^2\le H^2Pg$. Bernstein's upper-tail
inequality and $\sqrt{2ab}\le a+b/2$ give
\[
 P_ng\le Pg+\sqrt{2H^2(Pg)t/n}+2H^2t/(3n)
        \le2Pg+2H^2t/n
\]
except with probability $e^{-t}$. Cover each coefficient ball by a
Euclidean $H/n$-net. The number of pairs is at most
$(1+2nB/H)^{2d}$. Set
\[
 t=2d\log(1+4nB/H)+\log(8H/\delta)\le A_n.
\]
A union bound over pairs and layers costs at most $\delta/8$.
For any pair and a nearest pair of net points, uniformly on
$\norm{x}_2\le1$,
\[
 \abs{g_{w,\theta,h}(x)-g_{w',\theta',h}(x)}
 \le2H(\norm{w-w'}_2+\norm{\theta-\theta'}_2)
 \le4H^2/n.
\]
Extending the empirical term and twice the population term adds at most
$12H^2/n$. Since $A_n\ge1$, the total remainder is at most
$14H^2A_n/n$, and hence at most the stated $32H^2A_n/n$.
The event is proved for both full balls before any data-dependent
substitution is made.
\end{proof}

\subsection{Confidence for the original linear predictions}

\begin{lemma}[Raw confidence from clipped regression]
\label{stat:confidence}
Suppose the event of Lemma~\ref{stat:reference} holds. At a fixed layer,
let $F_h=Q_h^{\pi_{>h}}=\phi_h^{\top}\theta_h^*$ be a true suffix-policy
action-value function with $\norm{\theta_h^*}_2\le B$. Suppose the
feasible fit $\widehat\theta_h\in\cK_h$ has clipped prediction
$\bar q_h=\clip_{[0,L_h]}(\phi_h^{\top}\widehat\theta_h)$ and satisfies
$\E_{\mu_h}(\bar q_h-F_h)^2\le q$. Then
\begin{equation}
 \norm{\widehat\theta_h-\theta_h^*}_{V_h}^2
 \le71H^2A_n<R_n^2,
 \label{stat:raw-confidence}
\end{equation}
and, at every state--action pair,
\begin{equation}
 \abs{\phi_h(s,a)^{\top}\widehat\theta_h-F_h(s,a)}
 \le R_n\sqrt{\phi_h(s,a)^{\top}V_h^{-1}\phi_h(s,a)}=u_h(s,a).
 \label{stat:pointwise-confidence}
\end{equation}
The suffix policy and both coefficients may depend on the reference data.
\end{lemma}

\begin{proof}
The true value $F_h$ lies in $[0,L_h]$, so clipping its linear
representation does not change it. Since $q\le A_n/n$ and $H\ge1$,
Lemma~\ref{stat:reference} yields
\[
 \sum_{i=1}^n
  \abs{\clip_{[0,L_h]}(x_{ih}^{\top}\widehat\theta_h)
                         -x_{ih}^{\top}\theta_h^*}^2
 \le2nq+32H^2A_n\le34H^2A_n.
\]
For a reference prediction $t=x_{ih}^{\top}\widehat\theta_h$ and
$y=x_{ih}^{\top}\theta_h^*\in[0,L_h]$, feasibility implies
$\abs{t-\clip_{[0,L_h]}(t)}\le s_n$. Hence
\[
 \abs{t-y}^2\le2\abs{\clip_{[0,L_h]}(t)-y}^2+2s_n^2,
\]
and summing gives a raw squared-error sum at most
$68H^2A_n+2ns_n^2\le68H^2A_n+2H^2$.
The regularization contribution is at most
\[
 \lambda_0\norm{\widehat\theta_h-\theta_h^*}_2^2
 \le\frac{H^2}{4B^2}(2B)^2=H^2.
\]
Since $A_n\ge1$ and $R_n^2=128H^2A_n$, these bounds prove
\eqref{stat:raw-confidence}. Matrix Cauchy--Schwarz proves
\eqref{stat:pointwise-confidence}. The uniform reference event covers
the random true coefficient as well as the random fitted coefficient;
no independence from the reference block is used in this substitution.
\end{proof}

\subsection{Truncated leverage of the final design}

\begin{lemma}[Final-design truncated leverage]
\label{stat:leverage}
With probability at least $1-\delta/8$ over the reference block,
simultaneously at all layers,
\begin{equation}
 \E_{\mu_h}\min\{1,\phi_h(S,A)^{\top}V_h^{-1}\phi_h(S,A)\}
 \le J_n/n.
 \label{stat:leverage-event}
\end{equation}
Under all-policy density coverage, on this event every reachable state
marginal $\nu_h^S$ obeys
\begin{equation}
 \E_{\nu_h^S}\min\{H,U_h(S)\}^2
 \le v_n^2=CR_n^2J_n/n.
 \label{stat:uncertainty-moment}
\end{equation}
These statements hold for the final reference design and for
data-dependent choices of the reachable marginal. For the rounded
uncertainty $W_h=\kappa\lceil U_h/\kappa\rceil$, they also imply
\begin{equation}
 \E_{\nu_h^S}\min\{H,W_h(S)\}\le v_n+\kappa.
 \label{stat:rounded-uncertainty}
\end{equation}
\end{lemma}

\begin{proof}
Fix a layer and omit its subscript. Write
\[
 V_i=\lambda_0 I+\sum_{j=1}^ix_jx_j^{\top},\quad
 \mathcal F_i=\sigma(x_1,\ldots,x_i),\quad
 z_i=\min\{1,x_i^{\top}V_{i-1}^{-1}x_i\},\quad
 m_i=\E[z_i\mid\mathcal F_{i-1}].
\]
The inequality $\min\{1,t\}\le2\log(1+t)$, the matrix determinant
lemma, and $\tr(V_n)\le d\lambda_0+n$ give, pathwise,
\begin{equation}
 \sum_{i=1}^nz_i\le2\log\frac{\det V_n}{\det V_0}
 \le2d\log\left(1+\frac{n}{d\lambda_0}\right).
 \label{stat:pathwise-potential}
\end{equation}
The last inequality follows by the arithmetic--geometric mean inequality
for the eigenvalues of $V_n$.

Let $c_0=1-e^{-1}>1/2$. The chord bound
$e^{-z}\le1-c_0z$ on $[0,1]$ shows that
\[
 M_k=\exp\left(c_0\sum_{i=1}^km_i-\sum_{i=1}^kz_i\right)
\]
is a nonnegative supermartingale starting at one, because
\[
 \E[M_k\mid\mathcal F_{k-1}]
 \le M_{k-1}e^{c_0m_k}(1-c_0m_k)\le M_{k-1}.
\]
Markov's inequality and \eqref{stat:pathwise-potential} therefore imply,
with probability at least $1-\rho$,
\begin{equation}
 \sum_{i=1}^nm_i
 \le4d\log\left(1+\frac{n}{d\lambda_0}\right)+2\log(1/\rho).
 \label{stat:predictable-leverage}
\end{equation}

Define, for a fixed positive definite $V$,
\[
 \ell(V)=\E_{(S,A)\sim\mu_h}
       \min\{1,\phi_h(S,A)^{\top}V^{-1}\phi_h(S,A)\}.
\]
Independence gives $m_i=\ell(V_{i-1})$. On each complete reference
realization, $V_n\succeq V_{i-1}$, so
$\ell(V_n)\le\ell(V_{i-1})=m_i$ for every $i$. This is a pathwise
comparison of functions integrated against a fresh observation; it does
not require $V_n$ to be $\mathcal F_{i-1}$-measurable. Average over $i$
and use \eqref{stat:predictable-leverage}. Taking
$\rho=\delta/(8H)$ and a union bound over layers proves
\eqref{stat:leverage-event}, since the ceilings and base-two logarithms
in $J_n$ dominate the resulting natural-logarithm bound.

Now fix a reference realization on this event and a Markov prefix
producing $\nu_h^S$. At layer $h$, choose an action maximizing $u_h$,
and complete the policy arbitrarily. This is an admissible Markov policy
even when chosen using the data. Its state--action occupancy is bounded
by $C\mu_h$. Because $R_n^2\ge H^2$,
\[
 \min\{H^2,R_n^2t\}\le R_n^2\min\{1,t\}\qquad(t\ge0).
\]
Density coverage and \eqref{stat:leverage-event} consequently yield
\[
 \E_{\nu_h^S}\min\{H,U_h\}^2
 \le C\E_{\mu_h}
       \min\{H^2,R_n^2\phi_h^{\top}V_h^{-1}\phi_h\}
 \le CR_n^2J_n/n.
\]
Finally, $U_h\le W_h<U_h+\kappa$ and the truncation map is
one-Lipschitz. Cauchy--Schwarz proves
\eqref{stat:rounded-uncertainty}. Only truncated average uncertainty
is bounded; untruncated leverage can remain large on rare directions.
\end{proof}

\subsection{Conditional applications in the backward pass}

The two reference events fail with probability at most $\delta/8$ each.
On their intersection, proceed in the execution order $H,H-1,\ldots,1$.
At a stage whose later fits have succeeded, the previously constructed
single witness suffix makes the current label mean a true action-value
function. The reference block and later fitting blocks are already
exposed; the current fitting block remains independent. Lemma~\ref{stat:fast}
therefore has conditional failure probability at most $\delta/(8H)$.
Its success and Lemma~\ref{stat:confidence} provide the pointwise bound
needed to extend that same witness by the interval construction in the
main proof. The new stage does not alter the previously fixed suffix.

Partition a regression failure according to its first occurrence in this
execution order. The event of prior success is measurable before the
current block, so conditional expectations and a union bound over the
$H$ possible first failures give total probability at most $\delta/8$.
Together with the two reference events, all statistical guarantees used
by the backward induction hold with probability at least $1-3\delta/8$.
This argument neither assumes that a target remains realizable after a
failed future fit nor conditions fresh noise on a future success event.

\section{Deterministic budgets and finite-bit implementation}
\label{app:complexity}

This appendix gives an explicit reference-sample budget and verifies the
computational implementation. Throughout, $H,d,B,C\geq1$,
$0<\eps\leq1$, and $0<\delta<1$. Logarithms without a subscript are
natural. The algorithm solves exactly $H$ convex regressions in a single
backward pass.

\subsection{Selecting the reference sample before observing data}
\label{comp:budget}

Recall
\begin{align}
 \lambda_0&=\frac{H^2}{4B^2},\notag\\
 A_n&=1+2d\left\lceil\log_2\left(1+\frac{4nB}{H}\right)\right\rceil
       +\left\lceil\log_2\frac{8H}{\delta}\right\rceil,\notag\\
 J_n&=4d\left\lceil\log_2\left(1+\frac{n}{d\lambda_0}\right)\right\rceil
       +2\left\lceil\log_2\frac{8H}{\delta}\right\rceil,\label{comp:eq:log-budgets}\\
 R_n^2&=128H^2A_n,&
 v_n^2&=\frac{CR_n^2J_n}{n},&
 v_*^2&=\frac{\eps^2}{64H^2}.\notag
\end{align}
Starting at $n=1$, double $n$ until $v_n^2\leq v_*^2$. This search uses
only supplied parameters; no observations are requested until it ends.

\begin{proposition}[Explicit sample budget]
Set
\begin{align*}
 a&=A_1+4d,& b&=J_1+8d,&
 M&=\frac{128CH^2ab}{v_*^2}=\frac{8192CH^4ab}{\eps^2},\\
 s&=1+\lceil\log_2(2+M)\rceil,&
 n_*&=\lceil256Ms^2\rceil.
\end{align*}
The doubling search stops with $n\leq2n_*$. With
\[
 q=\min\left\{1,\frac{A_n}{n}\right\},\qquad
 s_n=\frac{H}{\lceil\sqrt n\rceil},
\]
use the integer logarithm budget and fitting-block size of the algorithm:
\begin{align}
 F_n&=d\left\lceil\log_2\left(1+\frac{128BH}{q}\right)\right\rceil
             +\left\lceil\log_2\frac{8H}{\delta}\right\rceil,\notag\\
 m&=\left\lceil\frac{64H^2F_n}{q}\right\rceil.
 \label{comp:eq:fit-budget}
\end{align}
These rounded binary logarithms dominate the natural-logarithm
regression bound and change its size by only a universal constant factor.
Then $N=n+Hm$ complete trajectories suffice. This is polynomial in
$H,d,C,\eps^{-1},\log(2+B),\log(1/\delta)$. Holding
$H,d,C,B,\delta$ fixed gives $N=\widetilde O(\eps^{-2})$.
\end{proposition}

\begin{proof}
For every integer $n\geq1$,
\[
 A_n\leq a(1+\log n),\qquad J_n\leq b(1+\log n).
\]
Indeed, $1+nx\leq n(1+x)$ and
$\lceil u+v\rceil\leq\lceil u\rceil+v+1$ bound the respective
increases over $A_1,J_1$ by $2d(1+\log_2n)$ and
$4d(1+\log_2n)$; the displayed $a,b$ dominate these bounds.
Since $M\geq1$, we have $n_*\leq257Ms^2$, $\log M\leq s$, and
$\log s\leq s-1$. Hence
\[
 1+\log n_*\leq3s+\log257-1<16s,
 \qquad n_*\geq M(1+\log n_*)^2.
\]
Thus $n_*$ satisfies the sufficient inequality for $v_{n_*}^2\leq v_*^2$.
The function $n/(1+\log n)^2$ is increasing for $n>e$, and $n_*>e$.
The first power of two at least $n_*$ is at most $2n_*$ and satisfies
the same sufficient inequality. This proves termination of the
deterministic search.

The quantities $a,b$ depend on $B$ only through logarithms, including
$\log(1+n/(d\lambda_0))$ with $\lambda_0=H^2/(4B^2)$.
Consequently $n_*$ and $n$ have polynomial dependence on the listed
parameters and only logarithmic dependence on $B$. Moreover,
$q^{-1}=\max\{1,n/A_n\}\leq n$; the remaining occurrence of $B$ in
\eqref{comp:eq:fit-budget} is inside a logarithm. This proves the total
budget assertion. At fixed $H,d,C,B,\delta$, the explicit bound gives
$n=\widetilde O(\eps^{-2})$, and the same is true of $m$ and $N$.
This statement concerns the accuracy exponent; it does not assert
optimal powers of the horizon, dimension, or coverage coefficient.
\end{proof}

One reference block supplies all layer designs. The uniform reference
square-error event and the truncated-leverage event may use the same
observations; their union bound does not require independence. Each has
failure probability at most $\delta/8$. With
\eqref{comp:eq:fit-budget}, each fresh fit has conditional failure
probability at most $\delta/(8H)$ on successful later-layer histories.
Summing probabilities of the first failed fit, together with the two
reference events, gives a total of at most $3\delta/8<\delta$.
No validation observations or additional optimization failure budget are
needed. Bounded clipped queries keep all labels in $[0,H-h+1]$ even
on histories where a statistical event fails, so the deterministic
computational bounds below hold on every outcome.

\subsection{Exact rational computation and rounded confidence queries}
\label{comp:bits}

For the bit-complexity statement, numerical parameters $B,C,\eps,\delta$,
observed rewards, and every evaluated feature coordinate are assumed
rational, with numerator and denominator encoding lengths at most
$\ell$. Feature evaluation is exact, $\norm{\phi}\leq1$, and rewards
belong to $[0,1]$. States and actions have finite representations. Costs
of reading states, enumerating actions, and evaluating features are
included separately. Unknown transition probabilities and the unknown
value coefficients need not be rational or supplied. These assumptions specify the finite-input model used for bit complexity.

We establish polynomial numerical training work in
\begin{equation}
 H,d,C,\eps^{-1},\log(2+B),\log(1/\delta),|\cA|,\ell,
 \label{comp:eq:bit-parameters}
\end{equation}
as well as polynomial output length and deployment cost. The arithmetic
version uses the same explicit separation and confidence computations in
the stated real-arithmetic model.

\paragraph{Parameters, design matrices, and upward-rounded uncertainty.}
For rational $x>0$, $\lceil\log_2x\rceil$ is computable by integer bit
lengths and a comparison with a power of two. Integer square roots give
$\lceil\sqrt n\rceil$. Thus all parameters, the doubling test, and the
integer sample counts in \eqref{comp:eq:fit-budget} have polynomial
rational encoding length. No exact natural-logarithm ceiling is required.

The positive definite reference matrix
\[
 V_h=\lambda_0I+\sum_{i=1}^n x_{ih}x_{ih}^{\top}
\]
and its inverse can be computed exactly by rational linear algebra.
Fraction-free elimination and determinant bounds give polynomial bit
lengths and computation time in $n,d,\ell$ and the encoding length of
$\lambda_0$. Small numerical eigenvalues do not replace these encoding
lengths by their reciprocals.

To keep the labels rational while preserving the confidence guarantee,
put $\kappa=\eps/(64H)$ and define
\begin{align}
 t_h(s)&=R_n^2\max_{a\in\cA}
              \phi_h(s,a)^{\top}V_h^{-1}\phi_h(s,a),\notag\\
 k_h(s)&=\min\{k\in\mathbb Z_{\geq0}:(\kappa k)^2\geq t_h(s)\},&
 W_h(s)&=\kappa k_h(s).
 \label{comp:eq:upward-grid}
\end{align}
The conceptual uncertainty is $U_h(s)=\sqrt{t_h(s)}$, and
\begin{equation}
 U_h(s)\leq W_h(s)<U_h(s)+\kappa.
 \label{comp:eq:upward-error}
\end{equation}
Every comparison in \eqref{comp:eq:upward-grid} is rational. In addition,
\[
 t_h(s)\leq R_n^2/\lambda_0=512B^2A_n,
 \qquad U_h(s)\leq32BA_n.
\]
One can therefore find $k_h(s)$ by integer binary search between zero
and $\lceil32BA_n/\kappa\rceil$. Its number of comparisons and the
integer's encoding length are
$O(1+\log(2+B)+\log A_n+\log(1/\kappa))$. Thus the search cost is logarithmic in the coefficient radius.

Use this same $W_h$ both in the decision condition
\[
 \max_a\widehat q_h(s,a)-\min_a\widehat q_h(s,a)\geq2W_h(s)
\]
and in the query
$v_h(s)=\clip_{[0,L_h]}(\max_a\widehat q_h(s,a)-W_h(s))$.
Suppose the raw confidence event holds for the current witness suffix,
so $|\widehat q_h(s,a)-Q_h(s,a)|\leq U_h(s)\leq W_h(s)$.
At a non-skipped state the rational un-clipped query obeys
\[
 \min_a Q_h(s,a)
 \leq\min_a\widehat q_h(s,a)+W_h(s)
 \leq\max_a\widehat q_h(s,a)-W_h(s)
 \leq\max_aQ_h(s,a).
\]
Thus it is already in $[0,L_h]$ and is exactly realizable by an action
mixture. Upward rounding preserves the exact witness construction and
does not introduce reward-model misspecification. The stopping decision
and query are both made with exact rational comparisons and arithmetic.

For the value bound, truncation is one-Lipschitz, so every relevant
state occupancy satisfies
\[
 \E\min\{H,W_h(S)\}
 \leq\E\min\{H,U_h(S)\}+\kappa\leq v_n+\kappa.
\]
The witness and greedy-action advantages are bounded respectively by
$4W_h$ and $2W_h$, and each is also bounded by $H$. Hence the
performance argument applies with truncated uncertainty and gives
\begin{equation}
 V^*-V^{\widehat\pi}\leq6H(v_n+\kappa)
 \leq\frac{3\eps}{4}+\frac{3\eps}{32}
 =\frac{27\eps}{32}<\eps.
 \label{comp:eq:rounded-performance}
\end{equation}

\paragraph{Regression geometry and exact feasible weak optimization.}
For a fitting block $(x_j,Y_j)_{j=1}^m$, the program is
\begin{align}
 F(w)&=\frac1m\sum_{j=1}^m(x_j^{\top}w-Y_j)^2,\notag\\
 \cK_h&=\{w:\norm{w}\leq B,\ -s_n\leq x_{ih}^{\top}w
                   \leq L_h+s_n\ (1\leq i\leq n)\}.
 \label{comp:eq:program}
\end{align}
It minimizes raw squared loss. Define $r=\min\{B,s_n\}$ and
$G=2(B+H)$. On every history,
\begin{equation}
 B_2(0,r)\subseteq\cK_h\subseteq B_2(0,B),\qquad
 0\leq F(w)\leq(B+H)^2,\qquad
 \norm{\nabla F(w)}\leq G\quad(\norm{w}\leq B).
 \label{comp:eq:geometry}
\end{equation}

\begin{lemma}[An exactly feasible approximate minimizer]
For rational data in \eqref{comp:eq:program} and rational $0<e\leq1$,
one can compute a rational $w_0\in\cK_h$ with objective gap at most
$e/2$ in time polynomial in the data encoding length, $m,n,d$, and
$\log(2+B),\log(1+1/r),\log(1/e)$. The output satisfies the original
constraints exactly.
\end{lemma}

\begin{proof}
For rational $t$, consider
$\cK_h(t)=\{w\in\cK_h:F(w)\leq t\}$. Membership of a rational
query point $x$ is decidable exactly. A violated slab directly provides
a rational separating inequality. If $\norm{x}^2>B^2$, then
\[
 x^{\top}y\leq(\norm{x}^2+B^2)/2
\]
contains the entire radius-$B$ ball and strictly excludes $x$. If
$x\in\cK_h$ but $F(x)>t$, convexity supplies
\[
 \nabla F(x)^{\top}y
 \leq\nabla F(x)^{\top}x+t-F(x).
\]
This is a rational strict separation. A zero gradient with $F(x)>t$
certifies an empty sublevel. These oracles have polynomial encoding cost.

Let $w_*$ minimize $F$ on $\cK_h$, and write $F_*=F(w_*)$. Set
\[
 \tau=e/32,\qquad
 \eta=\min\{1/2,\tau/[G(B+r)]\},\qquad \rho=\eta r.
\]
By convexity of the feasible set and the gradient bound,
\[
 B_2((1-\eta)w_*,\rho)\subseteq\cK_h,
 \qquad F(y)\leq F_*+G\eta(B+r)\leq F_*+\tau
\]
on that ball. Thus every level $t\geq F_*+\tau$ has a radius-$\rho$
ball inside its feasible sublevel, with $\log(B/\rho)$ bounded by the
logarithmic scales in the lemma.

Use the rounded rational ellipsoid method with outer radius $B$ and
this prescribed inner radius \cite[Chapters~4--5]{gls1988}. With the exact
separation oracle, it either queries an exactly feasible rational point
or produces an enclosing ellipsoid with volume smaller than a
radius-$\rho$ ball. Rounded updates preserve containment and volume
reduction with polynomial encoding lengths. A returned point is accepted
only after exact membership checks. The second outcome certifies the
absence of such an inner ball, not necessarily emptiness of the sublevel.
In dimension one, interval bisection gives the same alternative.

To optimize, start with $l=0$, $u=F(0)\leq H^2$, and feasible incumbent
$0$. Query $t=(l+u)/2$. On success, store the feasible point and put
$u=t$. On a no-inner-ball answer, the thickness argument implies
$t<F_*+\tau$; put $l=\max\{l,t-\tau\}$. The invariant
$l\leq F_*\leq u$ holds, and the incumbent has objective at most $u$.
The interval length decreases to at most half its previous length plus
$\tau$ at each step. After $O(1+\log((1+H^2)/e))$ steps it is at most
$e/4$. The incumbent therefore satisfies the claimed bound. The levels
use only midpoint operations and subtraction of fixed $\tau$, so their
bit lengths grow linearly with the number of bisection steps; long
solver-generated objective fractions are not fed back into the levels.
This gives an exactly feasible point with an additive objective guarantee.

The usual real-arithmetic ellipsoid analysis \cite[Theorem~5.2.1]{bental2001} likewise
uses only polynomially many separation and arithmetic operations in the
dimension and logarithmic scales. The rational rounded implementation
above supplies the additional finite-bit guarantee.
\end{proof}

\paragraph{A common feasible coefficient grid.}
Coordinatewise rounding toward zero need not preserve the signed slab
constraints. Instead set $e=q/8$ and, for the preceding feasible $w_0$,
define
\[
 \xi=\min\{1/2,e/[8G(B+r)]\},\qquad w_1=(1-\xi)w_0,
 \qquad
 p=\max\left\{0,\left\lceil\log_2\frac{2d}{\xi r}\right\rceil\right\}.
\]
Convexity gives $B_2(w_1,\xi r)\subseteq\cK_h$. Round $w_1$ to the
nearest coordinatewise multiple of $2^{-p}$, obtaining $\widetilde w$.
With fixed tie breaking,
\[
 \norm{\widetilde w-w_1}\leq\frac{\sqrt d}{2}2^{-p}
 \leq\frac{\xi r}{2},\qquad
 \norm{\widetilde w-w_0}\leq\xi(B+r/2).
\]
The rounded vector is exactly feasible, and its additional loss is at
most $G\xi(B+r/2)\leq e/8$. Its total objective gap is therefore less
than $e=q/8$, including rounding. In particular, a ball of radius
$\xi r/2$ around the rounded vector is still feasible.

The precision is fixed before fitting and satisfies
\[
 p=O\bigl(1+\log d+\log(2+B)+\log(2+H)+\log(1+1/r)+\log(1/e)\bigr).
\]
Every fitted vector used in a query or the output has this grid
representation, with coordinate encoding length $O(p+\log(2+B))$.
Internal solver fractions are not propagated as coefficients to earlier
layers. Every confidence radius is an integer multiple of the same
rational $\kappa$, with a multiplier of polynomial bit length as in
\eqref{comp:eq:upward-grid}.

\paragraph{Label lengths, total work, and deployment.}
A dot product of a grid coefficient and an $\ell$-bit feature vector
has polynomial encoding length. The quantity $W_h$ has encoding length
bounded by the encoding length of $\kappa$ plus
$O(1+\log(2+B)+\log A_n+\log(1/\kappa))$. Taking a maximum, subtracting
$W_h$, and clipping thus gives a rational $v_h$ with a uniform
polynomial bit bound. The skip variables are exactly computed binary
values. The recursion
\[
 Y_H=R_H,\qquad
 Y_t=R_t+(1-\omega_{t+1})v_{t+1}+\omega_{t+1}Y_{t+1}
\]
adds at most $H$ input rewards and one terminal query along a trajectory.
For example, all label lengths are bounded by
\[
 O\!\left(Hd\left[\ell+p+\log(2+B)+\log A_n+
                \log(1/\kappa)+\log(2+Hd)\right]\right).
\]
Summing rationals with different input denominators increases their bit
lengths by at most the sum of the input lengths. Therefore the empirical
quadratic coefficients, objective evaluations, gradients, and separation
cuts have polynomial bit length. The label bound is uniform over layers;
it depends on the common coefficient and confidence grids, not
recursively on the previous solver's internal precision. There is no
exponential bit-length growth through the backward pass.

There are $H$ optimization calls and polynomially many data-processing,
finite-action, and linear-algebra operations. The sample budget, the
logarithmic optimization scales, and both grid precisions are
polynomially bounded in \eqref{comp:eq:bit-parameters}. This proves the
total numerical work bound on every history. The statistical proof is
applied to the actual rounded coefficients and upward-rounded queries:
its feasibility, optimization tolerance, and exact witness conditions
remain valid.

The returned greedy policy stores $H$ coefficient vectors of total
length $O(Hd[p+\log(2+B)])$ and a fixed tie-breaking convention. At a
deployed state it only enumerates the actions and maximizes the raw
linear predictions. If newly evaluated feature coordinates have at
most $\ell_{\rm new}$ bits, one decision requires $|\cA|$ feature
evaluations and
$|\cA|\operatorname{poly}(d,\ell_{\rm new},p,\log(2+B))$ bit operations.
Reference matrices, confidence queries, and training trajectories are
not needed at deployment. The bounds include feature-access costs and
the encoding lengths of the exact rational inputs.

\section{Covariance coverage and uniform approximation error}
\label{sw:appendix}

The same interval-query construction admits a different statistical
implementation under all-policy covariance coverage. This variant retains
exactly $H$ backward regressions and permits uniformly approximate all-policy
realizability, at the cost of polynomial dependence on the numerical radius
$B$. The witness and value arguments are those of
Lemmas~\ref{lem:witness} and~\ref{lem:performance}; the differences are the
regression domain, confidence construction, and coverage transfer.

\subsection{Assumptions, algorithm, and guarantee}
\label{sw:algorithm}

Use the same layered MDP, initial distribution $\zeta$, finite action set $\cA$, and
complete independent trajectories from one unknown Markov behavior policy
$b$. Let $\mu_h=d_h^b$, $L_h=H-h+1$, and
$\|\phi_h(s,a)\|_2\le1$. For this appendix, replace exact realizability and
occupancy coverage by the following conditions. The known bounds $B,C\ge1$
and $\bar\eta\ge0$ satisfy, for every deterministic Markov policy $\pi$ and every
stage $h$,
\begin{align}
 &\text{there exists }\theta_h^\pi\in\R^d,
 \quad\|\theta_h^\pi\|_2\le B,
 \quad\sup_{s,a}|Q_h^\pi(s,a)-\phi_h(s,a)^\top\theta_h^\pi|
       \le\bar\eta,\label{sw:approx}\\
 &\Sigma_h^\pi\preceq C\Sigma_h,
 \qquad
 \Sigma_h^\pi=\E_{d_h^\pi}[\phi_h\phi_h^\top],
 \quad\Sigma_h=\E_{\mu_h}[\phi_h\phi_h^\top].
 \label{sw:coverage}
\end{align}
By Lemma~\ref{lem:randomization}, these conditions extend to all randomized Markov policies with the same constants. They impose
neither Bellman completeness nor a positive lower bound on the eigenvalues of
$\Sigma_h$. The coverage condition concerns feature covariances only, not
arbitrary functions or state-action density ratios. Values at points of zero
behavior probability are defined by forcing the initial state-action pair and
then using the same Markov kernels.

Fix $0<\eps\le1$, $0<\delta<1$, and a valid approximation bound satisfying
\begin{equation}
 0\le\bar\eta\le\frac{\eps}{64H\sqrt{Cd}}.
 \label{sw:eta}
\end{equation}
Set
\begin{align}
 r&=\frac{\eps}{8H\sqrt{Cd}},&
 \tau&=\frac r8,&
 \lambda&=\frac{r^2}{48B^2},&
 \rho&=\frac{\tau^2}{256B^2},\label{sw:parameters}\\
 \ell_D&=\log\frac{6dH}{\delta},&
 \ell_N&=d\log5+\log\frac{6H}{\delta}.&&&
 \label{sw:logs}
\end{align}
All parameters are chosen before data are inspected. Allocate one covariance
block and $H$ mutually independent regression blocks, with sizes
\begin{align}
 n_{\rm cov}&=\left\lceil\frac{28}{3\lambda}\ell_D\right\rceil,
 \label{sw:ncov}\\
 n_h&=\left\lceil\max\left\{
       \frac{28}{3\rho}\ell_D,
       64L_h^2\tau^{-2}\ell_N\right\}\right\rceil,
       \qquad h=1,\ldots,H,\label{sw:nh}\\
 N_{\rm cov}&=n_{\rm cov}+\sum_{h=1}^Hn_h.
 \label{sw:N}
\end{align}
From the covariance block compute
\begin{equation}
 \widehat\Sigma_h=\frac1{n_{\rm cov}}\sum_{i=1}^{n_{\rm cov}}
       \phi_h(S_h^i,A_h^i)\phi_h(S_h^i,A_h^i)^\top,
 \qquad \Lambda_h=2(\widehat\Sigma_h+\lambda I),
 \label{sw:design}
\end{equation}
and freeze
\begin{equation}
 u_h(s,a)=r\sqrt{\phi_h(s,a)^\top\Lambda_h^{-1}\phi_h(s,a)},
 \quad U_h(s)=\max_a u_h(s,a),
 \quad W_h(s)=U_h(s)+\bar\eta.
 \label{sw:uncertainty}
\end{equation}

Process stages $H,H-1,\ldots,1$. At stage $h$, use its fresh block to construct
the same stopped-return labels as in the main algorithm: continue along the
logged behavior suffix at skipped states and substitute $v_t$ at the first
later non-skipped state. All future queries and stopping rules are frozen
before this block is used. Fit the raw linear predictor by squared-loss
minimization over $\{\theta:\|\theta\|_2\le B\}$, to additive empirical
objective error at most $\tau^2/8$. There are no reference-point constraints
in this variant. Write $\widehat q_h=\phi_h^\top\widehat\theta_h$, and freeze
\begin{align}
 \omega_h(s)=0
 &\quad\Longleftrightarrow\quad
 \max_a\widehat q_h(s,a)-\min_a\widehat q_h(s,a)\ge2W_h(s),
 \label{sw:skip}\\
 v_h(s)&=\operatorname{clip}_{[0,L_h]}
                   \left(\max_a\widehat q_h(s,a)-W_h(s)\right).
 \label{sw:query}
\end{align}
Otherwise $\omega_h(s)=1$. Return the raw greedy policy
$\widehat\pi_h(s)\in\mathop{\rm argmax}_a\widehat q_h(s,a)$ with fixed
measurable tie breaking. Neither the behavior probabilities nor the witness
policy is an algorithmic input.

Each regression can be implemented by projected gradient descent on the $B$
ball, starting at zero with step size $1/2$. Its empirical loss has Hessian
$2\widehat\Sigma\preceq2I$. If $\theta^*$ minimizes this loss, the projection
optimality condition and smoothness imply
\[
 \widehat{\mathcal L}(\theta_{k+1})-
       \widehat{\mathcal L}(\theta^*)
 \le\|\theta_k-\theta^*\|_2^2-
       \|\theta_{k+1}-\theta^*\|_2^2.
\]
The objective is nonincreasing, so summing proves an error bound $B^2/k$.
Thus $\lceil8B^2/\tau^2\rceil$ iterations suffice. Clipping guarantees
$0\le Y_h\le L_h$ on every trajectory, including unsuccessful statistical
histories, and the algorithm has a fixed finite running-time bound.

\begin{theorem}[Single-pass learning under covariance coverage]
\label{sw:theorem}
Under \eqref{sw:approx}--\eqref{sw:eta}, the preceding algorithm satisfies,
with probability at least $1-\delta$,
\begin{equation}
 V^*-V^{\widehat\pi}
 \le6H(r\sqrt{Cd}+\bar\eta)\le\frac{27\eps}{32}<\eps.
 \label{sw:performance}
\end{equation}
It uses exactly $H$ regressions and the trajectory budget
\eqref{sw:N}. For a universal constant $c$,
\begin{equation}
 N_{\rm cov}\le
 cCH^3d\eps^{-2}
 \left\{B^2\log\frac{6dH}{\delta}
      +H^2\left[d+\log\frac{6H}{\delta}\right]\right\}+H+1.
 \label{sw:sample-bound}
\end{equation}
The real-arithmetic operation count is polynomial in
$H,d,|\cA|,C,B,\eps^{-1},\log(1/\delta)$.
\end{theorem}

\begin{corollary}[Exact covariance-coverage variant]
\label{sw:exact}
Under exact all-policy linear realizability and \eqref{sw:coverage}, take
$\bar\eta=0$. The same sample bound and algorithm give
$V^*-V^{\widehat\pi}\le3\eps/4$ with probability at least $1-\delta$.
\end{corollary}

\begin{corollary}[Arbitrary uniform approximation error]\label{sw:arbitrary-eta}
Under \eqref{sw:approx} and \eqref{sw:coverage}, let $\bar\eta\ge0$ be any known valid approximation bound. For $0<\eps\le1$ and $0<\delta<1$, a modification of the covariance-coverage algorithm returns a deterministic Markov policy satisfying, with probability at least $1-\delta$,
\[
V^*-V^{\widehat\pi}
\le\min\left\{H,\frac{3\eps}{4}+(48\sqrt{Cd}+6)H\bar\eta\right\}
\le\frac{3\eps}{4}+54H\sqrt{Cd}\,\bar\eta.
\]
It uses at most $H$ regressions, no more trajectories than the upper bound in \eqref{sw:sample-bound}, and the same polynomial real-arithmetic complexity.
\end{corollary}
\begin{proof}
If $\bar\eta>1/8$, return any fixed deterministic policy. Its loss is at most $H$, and $(48\sqrt{Cd}+6)H\bar\eta>H$.
Otherwise, set
\[
r=\max\left\{\frac{\eps}{8H\sqrt{Cd}},8\bar\eta\right\}
\]
and keep all other definitions in \eqref{sw:parameters}--\eqref{sw:query}. Then $0<r\le1$ and $\bar\eta\le r/8=\tau$. The proof of Theorem~\ref{sw:theorem} below uses these inequalities to obtain the first bound in \eqref{sw:performance}. Since
$r\le\eps/(8H\sqrt{Cd})+8\bar\eta$,
this gives the claimed additive guarantee. Increasing $r$ decreases every block size and the required number of gradient iterations, so the original sample and computational upper bounds remain valid.
\end{proof}

\subsection{Relative covariance and fixed-design regression bounds}
\label{sw:regression-sec}

\begin{lemma}[Regularized relative covariance concentration]
\label{sw:relative}
Let $X_1,\ldots,X_n$ be independent copies of $X\in\R^d$ with
$\|X\|_2\le1$, and let $\Sigma=\E[XX^\top]$ and
$\widehat\Sigma=n^{-1}\sum_iX_iX_i^\top$. For $0<a\le1$,
\begin{equation}
 \Pbb\left\{
 \left\|(\Sigma+aI)^{-1/2}(\widehat\Sigma-\Sigma)
                      (\Sigma+aI)^{-1/2}\right\|_{\rm op}>\frac12
 \right\}\le2d\exp(-3na/28).
 \label{sw:cov-tail}
\end{equation}
On the complementary event,
\begin{equation}
 \tfrac12\Sigma-\tfrac a2I\preceq\widehat\Sigma
 \preceq\tfrac32\Sigma+\tfrac a2I,
 \qquad
 \Sigma\preceq2(\widehat\Sigma+aI)\preceq3\Sigma+3aI.
 \label{sw:cov-order}
\end{equation}
\end{lemma}

\begin{proof}
For $Z=(\Sigma+aI)^{-1/2}XX^\top(\Sigma+aI)^{-1/2}$, one has
$0\preceq Z\preceq a^{-1}I$ and $\E Z\preceq I$. Hence
$\|Z-\E Z\|_{\rm op}\le a^{-1}$, and
\[
 \E[(Z-\E Z)^2]
 =\E Z^2-(\E Z)^2\preceq\E Z^2
 \preceq a^{-1}\E Z\preceq a^{-1}I.
\]
Self-adjoint matrix Bernstein~\cite{tropp2015}, at threshold $n/2$ for the
sum of the centered $Z_i$, gives
\[
 2d\exp\left[-\frac{(n/2)^2}
                       {2(n/a+(n/2)/(3a))}\right]
 =2d\exp(-3na/28).
\]
Conjugating the two-sided deviation bound by $(\Sigma+aI)^{1/2}$ proves
the first pair of inequalities in \eqref{sw:cov-order}. Adding $aI$ and
multiplying by two proves the remaining pair.
\end{proof}

\begin{lemma}[Regression with bounded conditional-mean error]
\label{sw:regression}
Let $(X_i,Y_i)$ be independent copies of $(X,Y)$, where $\|X\|_2\le1$ and
$0\le Y\le L$. Suppose $m(X)=\E[Y\mid X]$ satisfies
$|m(X)-X^\top\theta^*|\le\bar\eta$ almost surely for a vector
$\|\theta^*\|_2\le B$. Use $\tau=r/8$, $\rho=\tau^2/(256B^2)\le1$,
and $\bar\eta\le r/8$. For $0<p<1$, assume
\begin{equation}
 n\ge\frac{28}{3\rho}\log\frac{2d}{p},\qquad
 n\ge64L^2\tau^{-2}\left[d\log5+\log\frac2p\right].
 \label{sw:regression-size}
\end{equation}
If $\widehat\theta$ is a $\tau^2/8$-approximate empirical squared-loss
minimizer over the $B$ ball, then with probability at least $1-2p$,
\begin{equation}
 \E[(X^\top(\widehat\theta-\theta^*))^2]
 \le\frac{833}{4096}r^2.
 \label{sw:833}
\end{equation}
\end{lemma}

\begin{proof}
Write $\mathbf X$ for the design matrix and $\mathcal U$ for its column space.
Decompose
\[
 Y_i-X_i^\top\theta^*=\xi_i+b_i,
 \qquad \xi_i=Y_i-m(X_i),\qquad
 b_i=m(X_i)-X_i^\top\theta^*.
\]
Conditional on all design vectors, the $\xi_i$ are independent, centered,
and have ranges of length at most $L$. Thus for any fixed unit vector
$v\in\mathcal U$, conditional Hoeffding gives
\[
 \Pbb\{|v^\top\xi|>t\mid\mathbf X\}
 \le2\exp(-2t^2/L^2).
\]
The unit sphere of the at-most-$d$-dimensional space $\mathcal U$ has a
$1/2$-net of size at most $5^d$. A vector in $\mathcal U$ has norm at most
twice the largest absolute inner product with a point of that net. A union
bound therefore gives, with conditional probability at least $1-p$,
\begin{equation}
 \|\operatorname{Proj}_{\mathcal U}\xi\|_2^2
 \le2L^2\left[d\log5+\log\frac2p\right].
 \label{sw:noise}
\end{equation}
If $\mathcal U=\{0\}$, this bound is immediate. The net is only an analytical
device. By \eqref{sw:regression-size},
\[
 a_0:=\frac{\|\operatorname{Proj}_{\mathcal U}\xi\|_2}{\sqrt n}
 \le\frac{\tau}{4\sqrt2},\qquad
 a:=\frac{\|\operatorname{Proj}_{\mathcal U}(\xi+b)\|_2}{\sqrt n}
 \le a_0+\bar\eta.
\]
In particular, the bounded mean error $b$ is not treated as centered noise.

Let $\Delta=\widehat\theta-\theta^*$ and
$z^2=n^{-1}\|\mathbf X\Delta\|_2^2$. Since $\theta^*$ is feasible and
$\mathbf X\Delta\in\mathcal U$, empirical near-optimality yields
\[
 z^2\le\frac2n\langle\mathbf X\Delta,\xi+b\rangle+\frac{\tau^2}{8}
 \le2az+\frac{\tau^2}{8}
 \le\frac{z^2}{2}+2a^2+\frac{\tau^2}{8}.
\]
Hence $z^2\le4a^2+\tau^2/4$. Separately, Lemma~\ref{sw:relative} with
$a=\rho$ and the first condition in \eqref{sw:regression-size} gives, with
probability at least $1-p$, the simultaneous inequality
\[
 \frac1n\sum_i(X_i^\top v)^2
 \ge\frac12\E[(X^\top v)^2]-\frac\rho2\|v\|_2^2
 \quad\text{for all }v\in\R^d.
\]
Applying it to $\Delta$ and using $\|\Delta\|_2\le2B$ gives
\begin{equation}
 \E[(X^\top\Delta)^2]\le2z^2+4\rho B^2.
 \label{sw:design-transfer}
\end{equation}
The two events need not be independent; their joint failure probability is at
most $2p$. On their intersection,
\[
 a_0\le\frac{r}{32\sqrt2}\le\frac r{32},\qquad
 a\le\frac{5r}{32},\qquad
 z^2\le4(5r/32)^2+\frac{r^2}{256}=\frac{13r^2}{128}.
\]
Finally $4\rho B^2=\tau^2/64=r^2/4096$, so
\eqref{sw:design-transfer} is at most $(832+1)r^2/4096$.
\end{proof}

\subsection{Confidence, successful histories, and value transfer}
\label{sw:proof}

Lemma~\ref{sw:relative}, applied with $a=\lambda$ to each stage of the
covariance block, gives an event $\mathcal E_{\rm cov}$ of probability at
least $1-\delta/3$ on which
\begin{equation}
 \Sigma_h\preceq\Lambda_h\preceq3\Sigma_h+3\lambda I
 \quad\text{for all }h.
 \label{sw:cov-event}
\end{equation}
The samples at each stage are independent across trajectories. Dependence
between stages of a trajectory does not affect the union bound.

Condition on this block and the already processed future regression blocks,
assuming that those stages have succeeded. Lemma~\ref{lem:witness} supplies
one fixed Markov suffix policy whose value equals each non-skipped query and
which follows $b$ at skipped states. Its stopped-return identity gives
\begin{equation}
 F_h(s,a):=\E[Y_h\mid\text{start at }(s,a)]
          =Q_h^{\pi_{>h}}(s,a).
 \label{sw:target}
\end{equation}
This identity concerns a true policy and is exact, even when its linear
representation is only approximate. By \eqref{sw:approx}, choose
$\theta_h^*$ of norm at most $B$ approximating this true $Q$ uniformly within
$\bar\eta$. For $X=\phi_h(S_h,A_h)$, the tower property gives
\begin{equation}
 \E[Y_h\mid X]-X^\top\theta_h^*
 =\E\big[F_h(S_h,A_h)-X^\top\theta_h^*\mid X\big].
 \label{sw:tower}
\end{equation}
Its absolute value is at most $\bar\eta$. This step allows the feature map
to be noninjective and supplies the conditional mean required in
Lemma~\ref{sw:regression}. The current label rule is fixed before its fresh
block is used. Its regression success event therefore yields
\eqref{sw:833} with $L=L_h$.

On that event, let $\Delta_h=\widehat\theta_h-\theta_h^*$. Since both
coefficients belong to the $B$ ball, \eqref{sw:cov-event} gives
\begin{align}
 \Delta_h^\top\Lambda_h\Delta_h
 &\le3\E_{\mu_h}[(\phi_h^\top\Delta_h)^2]
                  +3\lambda\|\Delta_h\|_2^2\notag\\
 &\le\frac{2499}{4096}r^2+12\lambda B^2
  =\frac{3523}{4096}r^2<r^2.
 \label{sw:3523}
\end{align}
Matrix Cauchy--Schwarz and the uniform representation error imply
\begin{equation}
 |\widehat q_h(s,a)-Q_h^{\pi_{>h}}(s,a)|
 \le u_h(s,a)+\bar\eta\le W_h(s)
 \quad\text{for all }(s,a).
 \label{sw:confidence}
\end{equation}
At a non-skipped state, the interval-query argument in
Lemma~\ref{lem:witness}, using $W_h$, now places
$\max_a\widehat q_h-W_h$ in the true action-value interval. Clipping leaves
it unchanged. A mixture of true minimizing and maximizing actions realizes
this value exactly; at a skipped state the extension uses $b_h$. Thus the
same suffix witness extends to stage $h$. No future policy is changed and no
future regression error is inserted into \eqref{sw:target}. The base case is
$Y_H=R_H$. At the end, a single full policy $\pi_{\rm w}$ satisfies
\eqref{sw:confidence} at every stage, with its own true $Q_h^{\pi_{\rm w}}$.

To justify the good-history conditioning, apply
Lemma~\ref{sw:regression} with $p=\delta/(3H)$. The budget \eqref{sw:nh}
satisfies its two sample conditions, so each current stage has conditional
failure probability at most $2\delta/(3H)$ whenever the covariance and future
stages have succeeded. Decompose failure according to the first unsuccessful
stage in the order $H,H-1,\ldots,1$. The tower property then gives
\begin{equation}
 \Pbb\{\text{any failure}\}
 \le\frac\delta3+\sum_{h=1}^H\frac{2\delta}{3H}=\delta.
 \label{sw:first-failure}
\end{equation}
No realizability statement is made for a target produced after a failed
future stage; bounded labels suffice to define the algorithm there.

It remains to transfer uncertainty under the weaker coverage condition.
For any reachable stage-$h$ state distribution $\nu_h^S$, retain a Markov
prefix producing it and at that stage choose an action maximizing
$\phi_h(s,a)^\top\Lambda_h^{-1}\phi_h(s,a)$. This defines a permitted
Markov policy $\kappa$, also when the choice depends on the frozen data.
Equations~\eqref{sw:coverage} and~\eqref{sw:cov-event} give
\begin{align}
 \E_{\nu_h^S}U_h(S)^2
 &=r^2\operatorname{tr}(\Lambda_h^{-1}\Sigma_h^\kappa)\notag\\
 &\le Cr^2\operatorname{tr}(\Lambda_h^{-1}\Sigma_h)
 \le Cdr^2.
 \label{sw:transfer}
\end{align}
Thus $\E_{\nu_h^S}W_h\le r\sqrt{Cd}+\bar\eta$. This calculation transfers
only quadratic feature forms, not arbitrary errors through covariance
coverage. Applying Lemma~\ref{lem:performance} with the confidence width
$W_h$ (and bounding $\min\{H,W_h\}$ by $W_h$) yields
\[
 V^*-V^{\widehat\pi}
 \le6H(r\sqrt{Cd}+\bar\eta)
 \le\frac{3\eps}{4}+\frac{3\eps}{32\sqrt{Cd}}
 \le\frac{27\eps}{32}.
\]
This proves the probability and value assertions of
Theorem~\ref{sw:theorem}; setting $\bar\eta=0$ gives
Corollary~\ref{sw:exact}.

\subsection{Budget and scope of the variant}
\label{sw:budget}

Using $\max(x,y)\le x+y$, rounding each block size upward, and
$\sum_hL_h^2\le H^3$, the explicit sample budget obeys
\begin{equation}
 \begin{split}
 N_{\rm cov}\le H+1+448B^2r^{-2}\ell_D
       +\tau^{-2}\left[
          \frac{7168}{3}HB^2\ell_D+64H^3\ell_N\right].
 \end{split}
 \label{sw:expanded-budget}
\end{equation}
Substitute $\tau^{-2}=64r^{-2}$ and $r^{-2}=64CH^2d\eps^{-2}$, and use
$\ell_N\le(\log5)[d+\log(6H/\delta)]$, to obtain
\eqref{sw:sample-bound}. No statistical union bound over actions is needed:
the coefficient confidence event already controls all actions. Enumeration
still enters the arithmetic cost.

Condition~\eqref{sw:eta} gives an $\eps$-optimal guarantee, while Corollary~\ref{sw:arbitrary-eta} covers arbitrary known uniform approximation bounds. The approximation bound is an assumption on all policies and state-action pairs. This variant has polynomial dependence on the numerical value of $B$ and the real-arithmetic guarantee stated in Theorem~\ref{sw:theorem}.

\subsection{A lower bound for the accuracy exponent}
\label{sw:lower-sec}

\begin{proposition}[Two-action precision lower bound]
\label{sw:lower}
For $0<\eps\le1/16$ and $0<\delta<1/4$, there are two exact-realizable
instances with $H=1$, $d=2$, $B=1$, and $C=2$ such that any learner returning
an $\eps$-optimal policy with probability at least $1-\delta$ on each
instance must use at least
\begin{equation}
 n\ge\frac{\log(1/(4\delta))}{32(\log 3)\eps^2}
 \label{sw:lower-bound}
\end{equation}
independent complete trajectories. Both instances satisfy occupancy coverage
as well as covariance coverage. The learner and its output policy may be
randomized.
\end{proposition}

\begin{proof}
There is one state and two actions with features $\phi(a_1)=e_1$ and
$\phi(a_2)=e_2$. The behavior policy is uniform. Instance $+$ has Bernoulli
reward means $(1/2+4\eps,1/2)$; instance $-$ exchanges these means.
The true action-value vector is policy independent and has norm at most
$\sqrt{(3/4)^2+(1/2)^2}<1$. Thus exact all-policy realizability holds with
$B=1$. Every action distribution is bounded by twice the uniform behavior
distribution. Equivalently, its diagonal feature covariance is bounded above
by $I=2\Sigma$, proving both coverage claims with $C=2$.

An $\eps$-optimal output in instance $+$ assigns at least $3/4$ probability
to $a_1$, while an $\eps$-optimal output in instance $-$ assigns at most
$1/4$ probability to it. Comparing the output probability of $a_1$ with
$1/2$ therefore defines a test whose two error probabilities are at most
$\delta$.

Let $a=4\eps\le1/4$, and let $\operatorname{kl}$ denote binary relative
entropy. The relative entropy between the laws of one behavior observation
in the two instances is
\[
 \tfrac12\operatorname{kl}(1/2+a,1/2)
 +\tfrac12\operatorname{kl}(1/2,1/2+a).
\]
This expression equals
\[
 \frac a2\log\frac{1+2a}{1-2a}
 \le 2(\log 3)a^2=32(\log 3)\eps^2.
\]
Indeed, $\operatorname{atanh}(x)/x$ is increasing for $0<x<1$, as follows from its power series, and $2a\le1/2$.
For $n$ independent trajectories the relative entropy is at most
$32(\log 3)n\eps^2$; adding the learner's independent random seed leaves this bound
unchanged. The Bretagnolle--Huber testing inequality
\cite[Theorem~14.2]{lattimore2020} gives
\[
 2\delta\ge\tfrac12\exp(-32(\log 3)n\eps^2).
\]
Rearranging proves \eqref{sw:lower-bound}.
\end{proof}

This classical lower bound establishes the necessity of the
$\eps^{-2}$ exponent for both coverage models. It does not establish optimal
dependence on $H,d,B$, or $C$.

\end{document}